%% file: main.tex
\documentclass[10pt,twocolumn,letterpaper]{article}

\usepackage{pkg/cvpr}
\usepackage{times}
\usepackage{epsfig}
\usepackage{graphicx}
\usepackage{amsmath}
\usepackage{amssymb}

\usepackage{algorithm}
\usepackage{algorithmic}
\usepackage{bbm}
\usepackage{booktabs}
\usepackage{multirow}

\usepackage[breaklinks=true,bookmarks=false]{hyperref}

\cvprfinalcopy 

\def\cvprPaperID{****} 

\newcommand{\methodnameshort}{metaKernel}
\def\mathbi#1{\textbf{\em #1}}

\begin{document}

\title{Efficient Differentiable Neural Architecture Search with Meta Kernels}

\author{
  Shoufa Chen$^{1}$\thanks{This work was done when Shoufa Chen was a research intern at Yitu Technology.}, Yunpeng Chen$^{2}$, Shuicheng Yan$^{2}$, Jiashi Feng$^{3}$ \\
  $^{1}$Huazhong University of Science and Technology\\
  $^{2}$Yitu Technology \\
  $^{3}$National University of Singapore \\
  {\small shoufachen66@gmail.com, 
yunpeng.chen@yitu-inc.com, shuicheng.yan@yitu-inc.com,
, elefjia@nus.edu.sg}
}

\maketitle


\input{tex/1_abstract.tex}


\input{tex/2_introduction.tex}

\input{tex/3_related_work.tex}

\input{tex/4_method.tex}

\input{tex/5_experiment.tex}

\input{tex/6_discussion.tex}

\input{tex/7_conclusion.tex}

{\small
\bibliographystyle{pkg/ieee_fullname}
\bibliography{egbib}
}
\end{document}

%% file: tex/1_abstract.tex
\begin{abstract}
The searching procedure of neural architecture search (NAS) is notoriously time consuming and cost prohibitive. To make the search space continuous, most existing gradient-based NAS methods relax the categorical choice of a particular operation to a softmax over all possible operations and calculate the weighted sum of multiple features, resulting in a large memory requirement and a huge computation burden. In this work, we propose an efficient and novel search strategy with \emph{meta kernels}. We directly encode the supernet from the perspective on convolution kernels and ``shrink'' multiple convolution kernel candidates into a single one before these candidates operate on the input feature. In this way, only a single feature is generated between two intermediate nodes. The memory for storing intermediate features and the resource budget for conducting convolution operations are both reduced remarkably. Despite high efficiency, our search strategy can search in a more fine-grained way than existing works and increases the capacity for representing possible networks. We demonstrate the effectiveness of our search strategy by conducting extensive experiments. Specifically, our method achieves 77.0\% top-1 accuracy on ImageNet benchmark dataset with merely 357M FLOPs, outperforming both EfficientNet and MobileNetV3 under same FLOPs constraints. Compared to models discovered by start-of-the-art NAS method, our method achieves a same (sometimes even better) performance, while faster by three orders of magnitude.
\end{abstract}

%% file: tex/2_introduction.tex
\begin{figure}
    \begin{center}
    \includegraphics[width=0.9\linewidth]{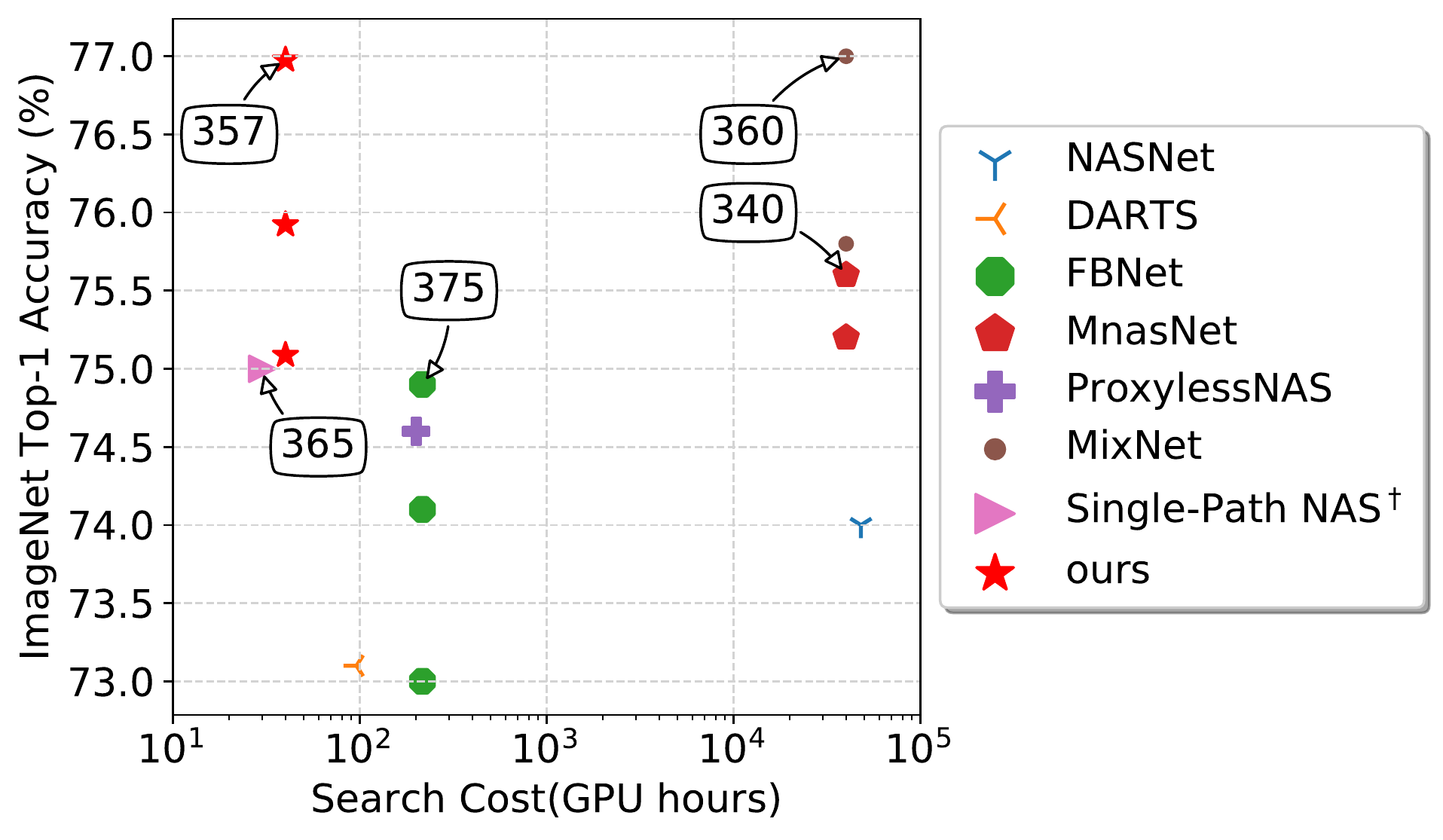}
    \end{center}
    
    \caption{Comparison of different NAS methods in terms of their search cost (GPU hours) in log scale and top-1 accuracy (\%) on ImageNet.     The numbers in the boxes denote the FLOPs (in million) of their searched models. $^\dagger$ denotes TPU hours. More detailed data are shown in Table~\ref{tab:imagenet-results}.}
    \label{fig:acc-GPUs}
\end{figure}

\section{Introduction}

Neural architecture search (NAS) has attracted lots of attention recently~\cite{zoph2016neural, gong2019autogan, liu2018darts, gao2019adversarialnas}.   However, its prohibitive time and computational resource cost is a remarkable problem that prevents its deployment   in many realistic scenarios. For example, the reinforcement learning (RL) based NAS method~\cite{zoph2018learning} requires 2000 GPU days and the evolutionary algorithm based method~\cite{real2019regularized} requires 3150 GPU days. Recent differentiable search methods, \eg, DARTS~\cite{liu2018darts}, reduce the  cost to some extent. However, DARTS still requires 96 GPU hours to search on a small proxy dataset CIFAR-10 and it is impractical to search on  large scale datasets like ImageNet~\cite{deng2009imagenet} directly.

\begin{figure*}[ht]
    \centering
    \includegraphics[width=0.9\textwidth, height=0.4\textwidth]{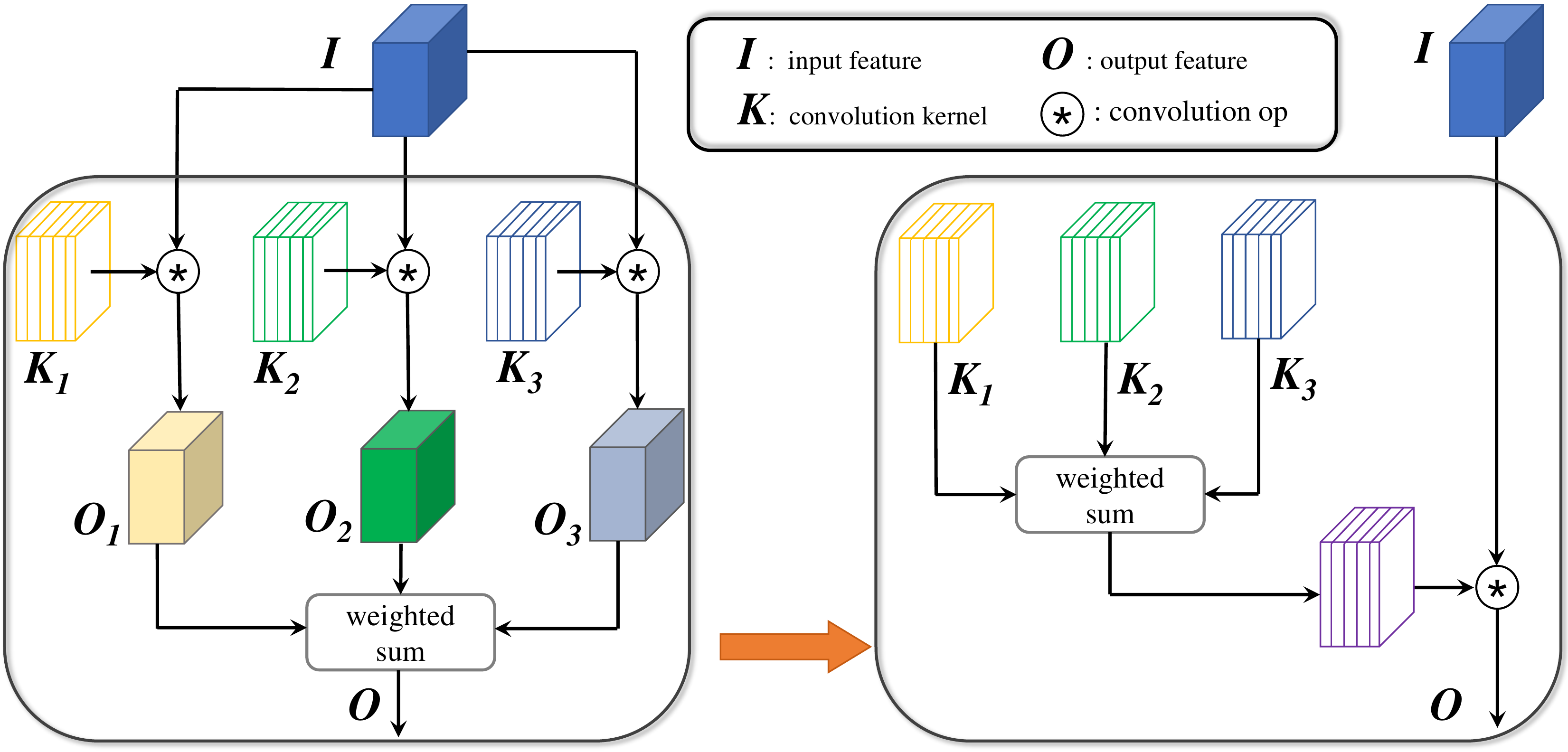}
    \caption{Illustration of the original DARTS~\cite{liu2018darts} supernet formulation (left) and our proposed efficient search formulation strategy (right). }
    \label{fig:big-pipeline}
\end{figure*}

The inefficiency of DARTS results from its strategy of aggregating multiple features generated by different candidate operations. {Following~\cite{liu2018darts, xie2019exploring}, we use directed acyclic graph (DAG) to represent the network work architecture and let the node/edge terminology denote the latent representation/candidate operation respectively in the network.} As illustrated in Figure~\ref{fig:big-pipeline} (left), multiple convolution candidates operate on the same input feature map and generate feature maps   respectively. The final output is the aggregation of these feature maps via a weighted sum. Conducting multiple convolution operations and storing the generated features bring the huge computation burden and memory cost.

In this work, we propose a novel and simple search method that reduces the cost of searching procedure significantly. The key idea of our method is to calculate the weighted sum of convolution kernels rather than output features, as illustrated in Figure~\ref{fig:big-pipeline} (right). We propose this strategy by exploiting the ``additivity'' of convolution, which has been  discussed in ACNet~\cite{ding2019acnet} recently. The ``additivity'' states that if several 2D kernels with \textit{compatible} sizes operate on the same input to produce outputs of the same resolution and their outputs are summed up, we can add up these kernels on the corresponding positions to obtain an equivalent kernel that will produce the same output~\cite{ding2019acnet}.  However, kernels with different shapes cannot be added up directly. To solve this problem, we propose a novel strategy, named as probabilistic kernel mask, which masks off the invalid area of a bigger kernel to present a smaller kernel, as illustrated in following section.

Based on the above ``additivity'' property, we develop a novel design of encoding the supernet that enables us to conduct the convolution operation on the input feature only once and get a single feature map between two intermediate nodes. Thus, the computation and memory cost can be reduced significantly compared to the previous search methods~\cite{liu2018darts, liang2019darts+, chen2019progressive}. {Our work suggests a new transition of searching for appropriate architectures, from evaluating feature maps to evaluating convolution kernels that generate feature maps.}

Because we take the convolution kernels as our direct search target, we can search in a more fine-grained way. Specifically, we can search for a convolution consisting of different kernel sizes.  Using different kernel sizes within a single convolution, we can increase the range of receptive fields, which means we can incorporate the multi-scale information within a single layer without changing the model's macro architecture. The idea of multi-scale representation has drawn great interest in the computer vision community and been applied to many vision tasks, such as classification~\cite{sun2019deep, huang2017multi, chen2019drop}, object detection~\cite{lin2017feature, singh2018sniper} and semantic segmentation~\cite{chen2017deeplab, liu2019auto}. Most of these works obtain multi-scale information by fusing multiple feature maps with different resolutions. In this work, we use the multi-scale information from the perspective on convolution kernel sizes.

This work makes contributions as following:
\begin{itemize}
 
    \item We propose a novel searching strategy that directly encodes the supernet from the perspective on convolution kernels and “shrink” multiple convolution kernel candidates into a single one before these candidates operate on the input feature. The memory  for storing intermediate features and the resource budget for conducting convolution operations are both reduced remarkably.
    
    \item Our search strategy is able to search in a more fine-grained way than previous methods and mix up multiple kernel sizes within a single convolution without the constraint in MixNet~\cite{tan2019mixconv}. The search space and the capacity for representing possible networks are significantly enlarged.
 
    \item Extensive experiments on both classification and object detection tasks are conducted. The results show the proposed searching method can discover new state-of-the-art light-weight CNNs while successfully reducing the searching cost by about three orders of magnitude than existing SOTA.
    \end{itemize}

%% file: tex/3_related_work.tex
\section{Related Work}

\paragraph{Efficient Search methods} Neural architecture search (NAS) has relieved substantial handcraft efforts for designing a neural network architecture and  has been explored in many computer vision tasks, such as classification~\cite{zoph2016neural, zoph2018learning, liu2018darts}, detection~\cite{ghiasi2019fpn, chen2019detnas}, semantic segmentation~\cite{liu2019auto, ghiasi2019fpn} and GAN~\cite{gong2019autogan}. However, the prohibitive cost for NAS is still a remarkable problem.  For example, the reinforcement learning (RL) based NAS method~\cite{zoph2018learning} requires 2000 GPU days and the evolutionary algorithm based method~\cite{real2019regularized} requires 3150 GPU days.  

The gradient-based methods relax a discrete architecture choice to a continuous search space, allowing search of the architecture using gradient descent~\cite{liu2018darts, wu2019fbnet, fang2019densely, dong2019searching}. Although gradient-based methods are more efficient than RL  and evolutionary based ones, the adopted  relaxation  still brings heavy computation and memory burden for calculating and storing the multiple features generated by all possible candidates. ProxylessNAS~\cite{cai2018proxylessnas} propose to binarize the architecture parameters and force only one path to be active during running, which reduces the required memory through requires GPU memory management. However, at least 200 GPU hours are still needed in~\cite{cai2018proxylessnas}.

Single-Path NAS~\cite{stamoulis2019single} is a differentiable search algorithm with only a single path between two intermediate nodes. It views the small kernel as the core of the large kernel. Single-Path NAS chooses a candidate based on the L2 norm of the convolution weight. Specifically, it formulates a condition function in which the L2 norm of the convolution weight is compared to a threshold that  controls the choice of   convolution kernels. Our proposed method is very different from Single-Path NAS. First, we directly use explicit architecture parameters to represent the importance of all candidates, while Single-Path NAS uses the comparison results of convolution weights and threshold value. Second, the complexity of the condition function used in Single-Path NAS increases linearly with the number of kernel candidates, while our method can be applicable to any number of candidates readily. Third, Single-Path NAS searches for a single kernel size within a convolution, rather than multiple kernel sizes as done in ours.

\paragraph{Multi-Scale Representation} Multi-scale representation has been widely explored in computer vision~\cite{lin2017feature, huang2017multi, chen2019drop, gao2019res2net}. Some works introduce multi-scale information from the perspective on macro architecture and design model architectures with multi-branch topology~\cite{huang2017multi, wang2019deep, sun2019deep}. Others propose to re-design the convolution operation~\cite{chen2019drop, gao2019res2net} and combine multi-scale information in a single convolutional layer, without modifying the macro architecture. 

In recent   MixConv~\cite{tan2019mixconv}, Tan \etal also proposed to use   similar convolutions. However, all the convolution candidates in their method are always split uniformly. Thus, the search space used in~\cite{tan2019mixconv} is limited due to such fixed allocation  of different kernels.
It is reasonable that the convolutional layers' preferences for kernel size differ across the network. So keeping a fixed ratio for the network at different depths would not achieve optimal performance. Nonetheless, it is not practicable to manually fine-tune an \textit{ad-hoc} ratio for a specific layer because of the non-trivial burden introduced by endless trial and error.
In this work, we remove the ``uniform partition'' constraint and search for every kernel independently, which means the search space is significantly enlarged and specifically, the search space used in MixNet~\cite{tan2019mixconv} is a subset of ours. Benefiting from no constraint, the convolution operations are mixed up more robustly and flexibly than MixNet~\cite{tan2019mixconv}.

%% file: tex/4_method.tex
\section{Method}
We start with preliminaries on the additivity property of convolutions~\cite{ding2019acnet}, which is the theoretical basis for our efficient search strategy. We then introduce the Probability Masks, which is designed for representing the supernet from the perspective on convolution kernels. Finally, in order to discover appropriate models with different computation resource budgets, we employ a resource-aware search objective function following~\cite{dong2019searching}.

\subsection{Additivity of Convolution}

Consider there are $N$ 2D convolutional kernels $\textbf{\textit{K}}^{(i)}$ that operate on the same input $\textbf{\textit{I}}$ separately. If these 2D kernels have the same stride and \textit{compatible} sizes, the sum of their outputs can be obtained in an equivalent way: adding up these $N$ kernels on the corresponding positions to formulate a single kernel, and then conducting convolution operation on the input with this generated single kernel to get the final output. Here \textit{compatible} means that the smaller kernel can be generated by slicing the larger kernel. For example, $3\times1, 1\times3$ kernels are \textit{compatible} with $3\times3$ ~\cite{ding2019acnet}.  Such  `additivity' of convolution can be formally represented as 
\begin{align}\label{eq:additivity}
    \sum_i^{N} \textbf{\textit{I}} \ast \textbf{\textit{K}}^{(i)} = \textbf{\textit{I}} \ast ( \textbf{\textit{K}}^{(1)} \oplus  \textbf{\textit{K}}^{(2)} \oplus  \cdots \oplus \textbf{\textit{K}}^{(N)} ),
\end{align}
where 
$\oplus$ denotes the element-wise addition  of the kernel parameters and $\sum$ denotes element-wise addition of the resulted features. 

To the best of our knowledge, this is the first work that introduce the additivity of convolution to the NAS fields. We experimentally show that using this property can help reduce the searching time remarkably.
 
\subsection{Meta Convolution Kernels} 
We propose an efficient search algorithm that can significantly improve   efficiency of the search process based on the additivity of convolution discussed above. {The key part of our search strategy is that   we use the weighted sum of kernels, rather than weighted sum of feature maps used in previous works~\cite{liu2018darts, chen2019progressive, dong2019searching}, to represent the aggregation of multiple outputs generated by all edges (candidates operations).
}

Let $\mathbb{K} = \{ \mathbi{K}^{(1)}_{(w_1, h_1)}, \mathbi{K}^{(2)}_{(w_2, h_2)}, \cdots, \mathbi{K}^{(N)}_{(w_{N}, h_{N})} \}$ denote a set of candidate kernels,  where $(w_i, h_i)$ represents the width and height of the $i$-th kernel, respectively.  We use the architecture parameters $\alpha = \{ \alpha^{(1)}, \alpha^{(2)}, \cdots, \alpha^{(N)} \}$ to encode the over-parameterized kernel at the search stage. The $\alpha^{(i)}$ represents the probability of selecting $\mathbi{K}^{(i)}_{(w_i, h_i)}$ as the   candidate, correspondingly.

\subsubsection{Continuous Relaxation and Reformulation} 

The previous gradient-based NAS methods~\cite{liu2018darts, chen2019progressive, dong2019searching} relax the categorical choice to a softmax one over  {multiple candidate operations}. It  can be formulated as:
\begin{align}\label{eq:origin-relax}
    \mathbi{O} = \sum_{i=1}^{N} \frac{\exp(\alpha^{(i)})}{\sum_j^{N} \exp(\alpha^{(j)})} \cdot (\mathbi{I} \ast \mathbi{K}^{(i)}),
\end{align}
where $\mathbi{O}$ denotes the output, a weighted sum of features from multiple operations. As stated above, multiple output features need to be calculated and stored between two nodes  and the weighted sum over these multiple features is taken as the final output of a node.

Based on additivity of convolution in Eq.~\eqref{eq:additivity}, we reformulate Eq.~\eqref{eq:origin-relax} as:
\begin{align}\label{eq:new-relax}
    \mathbi{O} = \mathbi{I} \ast \sum_{i=1}^{N}  \left( \frac{\exp(\alpha^{(i)})}{\sum_j^{N} \exp(\alpha^{(j)})} \cdot  \mathbi{K}^{(i)} \right),
\end{align}
where the outer $\sum_j^{N}$ denotes the element-wise addition operation of the kernel parameters on the corresponding position. Through such reformulation, we can combine multiple candidate kernels into a single one before they operate  on the features. Thus, we just need conduct the convolution operation once and generate a single output feature between two intermediate nodes, avoiding the intrinsic inefficient problems introduced by multi-paths.

\subsubsection{Candidate Kernel Formulation}\label{sec:mask-step} 
Now, we introduce  details of our search strategy. It consists of three steps. The first step is to determine the \emph{meta kernels}; the second step is to generate probabilistic masks over the meta kernels; and the third step is to sample all the candidate kernels from meta kernels by  the  probabilistic masks.

\paragraph{Step 1: Build meta kernels}  We first build a special kernel $\hat{\mathbi{K}}_{(\hat{w}, \hat{h})}$ with the shape of
\begin{align}
    \hat{w} = \max(w_i) \text{    and    } \hat{h} =\max(h_j), 
\end{align}\label{eq:matrix-shape}
where $i,j \in \{1,2,\cdots, N \}$. This implies that all kernels in the set $\mathbb{K}$ are compatible to $\hat{\mathbi{K}}$. We name the kernel $\Hat{\mathbi{K}}$ as \emph{meta kernels} because all of the candidates in the set $\mathbb{K}$ originate from it.  For example, for a candidate set of $3\times3, 5\times5, 7\times7$ kernels, the corresponding meta kernel has the shape of $7\times7$.

\begin{figure}[t]
    \centering
    \includegraphics[width=0.8\linewidth]{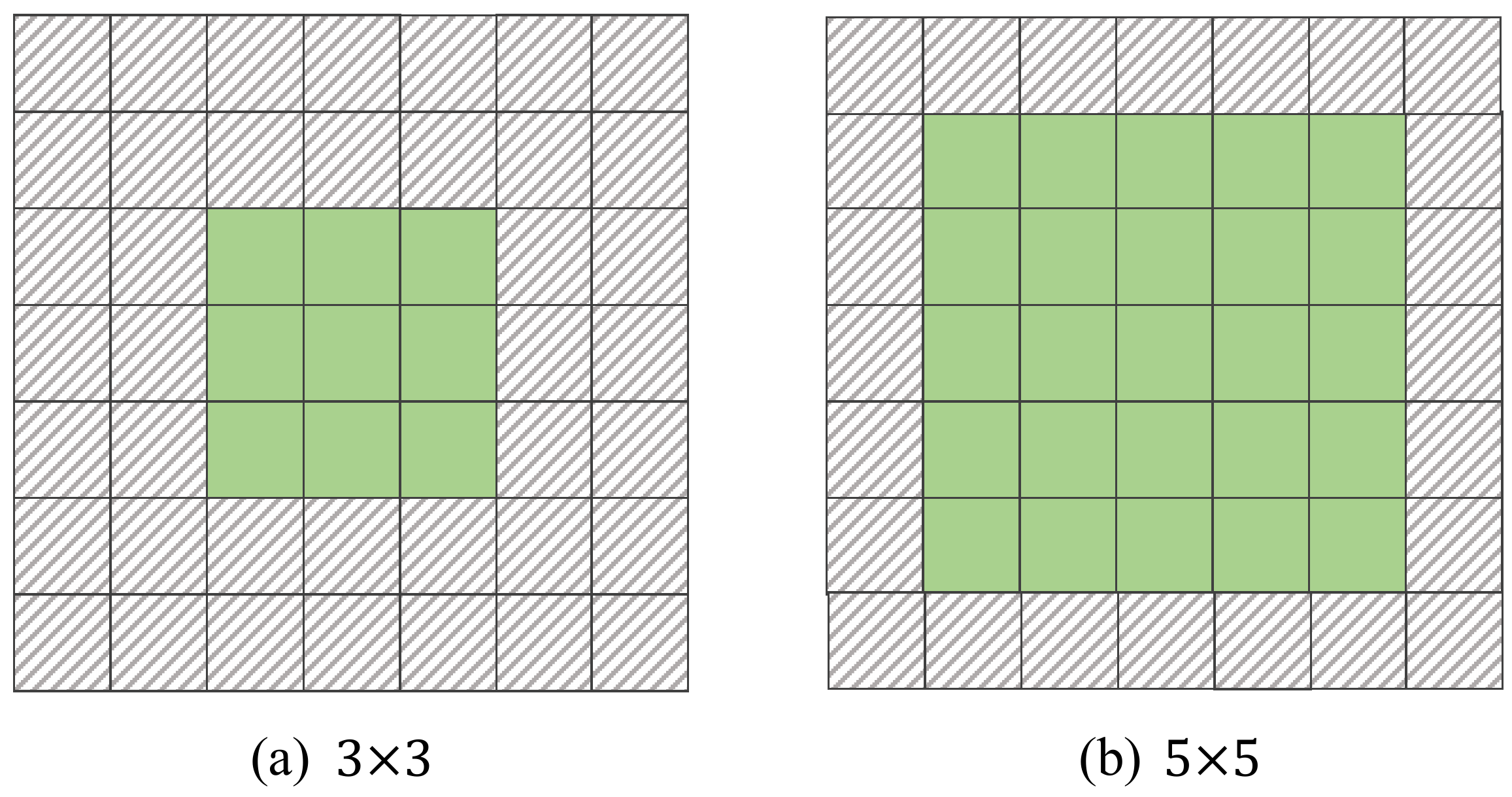}
    \caption{Illustration of two mask examples. The area with green color is the mapping the corresponding kernel in mask. We denote this mapping area as \textit{RoI} of $\mathbi{K}^{(i)}$. For a specific mask, all elements in the \textit{RoI} area are filled with the same probability value. Elements in diagonally hatched grey area are filled with \textit{zero}.  Best viewed in color.}
    \label{fig:RoI-mask}
\end{figure}

\paragraph{Step 2: Learn the probability mask} 
Given the kernel candidate set $\mathbb{K}$, there is another corresponding mask set, $\mathbb{M}=\{ M^{(1)}, M^{(2)}, \cdots, M^{(N)} \}$, which serves as the intermediary of over-parameterizing the candidate kernels with architecture parameters $\alpha$. Each mask $M^{(i)}$ has the same shape as the KermelMatrix, $\Hat{\mathbi{K}}$. The elements of $M^{(i)}$ are defined as:
\begin{align}\label{eq:mask-define}
    m^{(i)}_{(x,y)} = 
    \begin{cases}
    \mathcal{P}(\mathbi{K}^* = \mathbi{K}^{(i)}),  & (x,y) \in \textit{RoI of   \space} \mathbi{K}^{(i)} \\
    0 & \textit{otherwise}
    \end{cases}
\end{align}
where $\mathcal{P}(\mathbi{K}^* = \mathbi{K}^{(i)})$ is the sampling probability of $\mathbi{K}^{(i)}$ at search stage. And we define the \textit{RoI of} $\mathbi{K}^{(i)}$ as the mapping of the $\mathbi{K}^{(i)}$ in mask $M^{(i)}$, as illustrated in Figure~\ref{fig:RoI-mask}. The mapping area in the $M^{(i)}$ is determined following two principles: (1) The center of the \textit{RoI} is located at the center of $M^{(i)}$ and (2) the shape of the \textit{RoI} is same as its corresponding kernel candidates $\mathbi{K}^{(i)}$. Note that extra memory introduced by $\mathbb{M}$ is negligible compared to that introduced by feature maps from multi-paths, as used in previous works.

\paragraph{Step 3: Generate all the candidate kernels} Now, every candidate kernel, $\mathbi{K}^{(i)}$, can be generated by multiplying its corresponding mask $M^{(i)}$ and the KernelMaster $\Hat{\mathbi{K}}$. 
Based on the above formulation, we add an extra $M^{(0)}$ into the mask set $\mathbb{M}$, which serves as controlling the total number of the filters in a layer. We name $M^{(0)}$ as \textit{None} as all elements in $M^{(0)}$ are equal to zero. With the help of $M^{(0)}$, some redundant filters can be pruned at the search stage.

Note that in the above discussion, for the sake of simplicity, we take the search process of a single filter within a convolution layer as an example. However, it is easy to extend to all filters because every kernel is treated independently at the search stage.
Furthermore, benefiting from our fine-grained search strategy, the vanilla depthwise convolution and the mixed convolution proposed in~\cite{tan2019mixconv} are three special cases of our search space.

\subsection{Search with Cost-aware Objective}
In order to let our proposed method generate models adaptively under different circumstances, we incorporate the cost-aware constraint into our objective to formulate a multi-objective search algorithm. Formally, we use the FLOPs as the proxy of the computation consumption and the corresponding searching loss is defined as
\begin{align}\label{eq:flops-loss}
    \mathcal{L}_{FLOPs} = 
    \begin{cases}
    - \log \mathbb{E}[\mathcal{C}(\mathbb{A})]& \text{ if } \mathbb{E}[\mathcal{C}(\mathbb{A})] < T\cdot(1-\eta) \\
      \log \mathbb{E}[\mathcal{C}(\mathbb{A})] & \text{ if }  \mathbb{E}[\mathcal{C}(\mathbb{A})] > T\cdot(1+\eta) \\
      0  & \text{otherwise}
    \end{cases}
\end{align}
where $T$ is the computation cost budget, which can be adapted according to different needs. The function $\mathcal{C}$ counts the FLOPs of   a specific architecture  $\mathbb{A}$ sampled from the search space at the search stage. $\eta$ is a slack variable. As the FLOPs of a sampled network is a discrete value so it is reasonable to confine the FLOPs in a small range rather than a single point.  We regard FLOPs as our cost-aware supervision in this work and other metrics such as latency as used in \cite{wu2019fbnet, tan2019mnasnet,li2019partial} can replace FLOPs as the objective readily.

\subsection{Differentiable Search Algorithm}
With introducing the cost-aware loss $\mathcal{L}_{FLOPs}$, we search for the network architectures to minimize  the following  multi-objective loss:
\begin{align}\label{eq:final-objective}
    \mathcal{L}(a, w_a) = \mathcal{L}_{CE} + \lambda_{cost}\cdot \mathcal{L}_{\text{FLOPs}}
\end{align}
where $a$ represents an architecture in the search space and  $w_a$ denotes the convolution weights of the corresponding model. We adopt the differentiable search method to solve the problem of finding the optimal kernels.

The probability of sampling the $i$-th kernel candidate in the Eq.~\eqref{eq:mask-define} is computed  as
\begin{align}\label{eq:softmax-probability}
    \mathcal{P}(\mathbi{K}^* = \mathbi{K}^{(i)}) = \frac{\exp(\alpha^{(i)})}{\sum_{j=0}^{N} \exp(\alpha^{(j)})}
\end{align}

Instead of directly relaxing the categorical choice of a particular kernel to a softmax over all possible candidates as Eq.~\eqref{eq:new-relax},  we formulate the search stage as the sampling process, as done in~\cite{dong2019searching, wu2019fbnet}.

\begin{algorithm}[t]
	\caption{The \methodnameshort{} algorithm}
	\label{alg:main}
	\begin{algorithmic}
		\STATE {\bfseries Input:} Search space $\mathbb{K}$, FLOPs target T,  randomly initialized architecture parameters $\alpha$ and convolution kernel parameters $\mathbi{K}$, dataset $\mathbb{D}_{train}$ 
        \WHILE{not converge}
        \STATE 1. Generate all kernel candidates from the meta kernel $\Hat{\mathbi{K}}$ by means of the probability mask set $\mathbb{M}$
        \STATE 2. Aggregate multi-paths into a single one based on Eq.~\eqref{eq:new-relax}
        \STATE 3. Calculate $\mathcal{L}(a, w_a)$ based on Eq.~\eqref{eq:final-objective}
        \STATE 4. Update weights: $w \gets w - \partial_w \mathcal{L}$
        \STATE 5. Update probability parameter: $\alpha \gets \alpha - \partial_{\alpha} \mathcal{L}$
        \ENDWHILE
        \STATE Derive the final kernel combination from the learned $\alpha$.
        
	\end{algorithmic}
\end{algorithm}

Although the objective function is differentiable with respect to the weight of kernel $\mathbi{K}$, it is not differentiable to the architecture parameters $\alpha$ due to the sampling process. In order to sidestep this problem, we adopt the Gumbel Softmax function~\cite{maddison2016concrete, jang2016categorical}, as used in recently NAS related works~\cite{wu2019fbnet, cai2018proxylessnas, dong2019searching, xie2018snas}. The sampling probability in Eq.~\eqref{eq:softmax-probability} can be rewritten as
\begin{align}\label{eq:gumbel-probability}
    \mathcal{P}(\mathbi{K}^* = \mathbi{K}^{(i)}) = \frac{\exp{( (\log(\pi_i) + g_i)/\tau )}}{ \sum_j \exp{( (\log(\pi_j) + g_j)/\tau )}    }
\end{align}
where $g_j$ is sampled form the distribution of Gumbel (0,1), and $\pi_i$ is the class probabilities of categorical distribution calculated by Eq.~\eqref{eq:softmax-probability}.

After the searching process, we can derive the architecture from the architecture parameters $\alpha$. Our pipeline is summarized in Algorithm~\ref{alg:main}. We will show in the next experiment section that our proposed search algorithm costs orders of magnitude less search time than previous RL based NAS and gradient-based multi-paths NAS while achieving better performance.

%% file: tex/5_experiment.tex
\section{Experiments}

In this section, we aim to validate  effectiveness of our proposed search method. We first conduct ablation studies to investigate the effectiveness of mixing multiple kernel sizes without any constraints, which is more flexible than MixConv~\cite{tan2019mixconv}. Then, we compare our searched models with state-of-the-art , both manually designed and discovered by NAS methods. Besides, we further conduct object detection experiments to show the advantage of our models as a backbone feature  extractor.
    
\subsection{Implementation Details} 

We conduct experiments on the widely used  ImageNet~\cite{deng2009imagenet} benchmark.  
We use the normal data augmentation including random horizontal  flipping   with 0.5 probability, scaling hue/saturation/brightness, resizing and cropping, following~\cite{he2019bag}. We do not use the mixup~\cite{zhang2017mixup} or AutoAugment~\cite{cubuk2018autoaugment} for a fair comparison.  The models are trained for 250 epochs from scratch as done in \cite{liu2018darts, dong2019searching, chen2019progressive}. We train the models on 8 Nvidia 2080Ti GPUs with a total batch size 1024.  The learning rate is initialized as 0.65 and decayed to 0 at the end of the training stage, following the cosine rule. We use a weight decay of $3 \times 10^{-5}$. At the evaluation phase, we adopt the popular   settings, i.e. resizing the image into $256\times256$ and then center cropping a single $224\times224$ patch. We set $\lambda_{cost}$ as 2.0 and $\eta$ as 0.1.

\subsection{Ablation Study}
Because we search for an appropriate ratio of different kernel sizes within a single depthwise convolution, we conduct a series of ablation studies to demonstrate that our proposed method can achieve better FLOPs-Accuracy trade-off than both vanilla depthwise convolution and MixConv~\cite{tan2019mixconv} 
where multiple kernels are mixed up in a manually designed partition way. 

\subsubsection{Settings}
Following MixConv~\cite{tan2019mixconv}, we design three kinds of baseline settings to implement the depthwise convolution:
\renewcommand{\theenumi}{\Alph{enumi}}
\begin{enumerate}
    \item Single kernel size within a depthwise convolution.
    \item Multiple kernel sizes in an uniform partition way.
    \item Multiple kernel sizes in an exponential partition way.
\end{enumerate}
Note that the above three baseline models are three special cases of our search space  because our search algorithm aims to find the proper ratio of different kernels within a single convolution operation.

To perform apple-to-apple comparison, we reproduce all baseline methods under the same training/testing setting for internal ablation studies. Following MixConv~\cite{tan2019mixconv}, we conduct all experiments on the widely used MobileNetv1~\cite{howard2017mobilenets} networks.

For baseline A, we start with the original MobileNetv1 and then replace depthwise $3\times3$ convolution with $5\times5, 7\times7$ and $9\times9$ ones, respectively.  For baselines B and C, we adjust the number of kernel types from 1 to 6.  The kernel sizes increase from $3\times3$, with a step   of 2. For example, when the number of kernel types is 6, the corresponding kernel candidate set is $\{3\times3, 5\times5, 7\times7, \cdots, 13\times13\}$. The candidate sets on which we conduct our search method are the same as baselines B and C for  fair comparison.

\begin{figure}
\begin{center}
\includegraphics[width=0.9\linewidth]{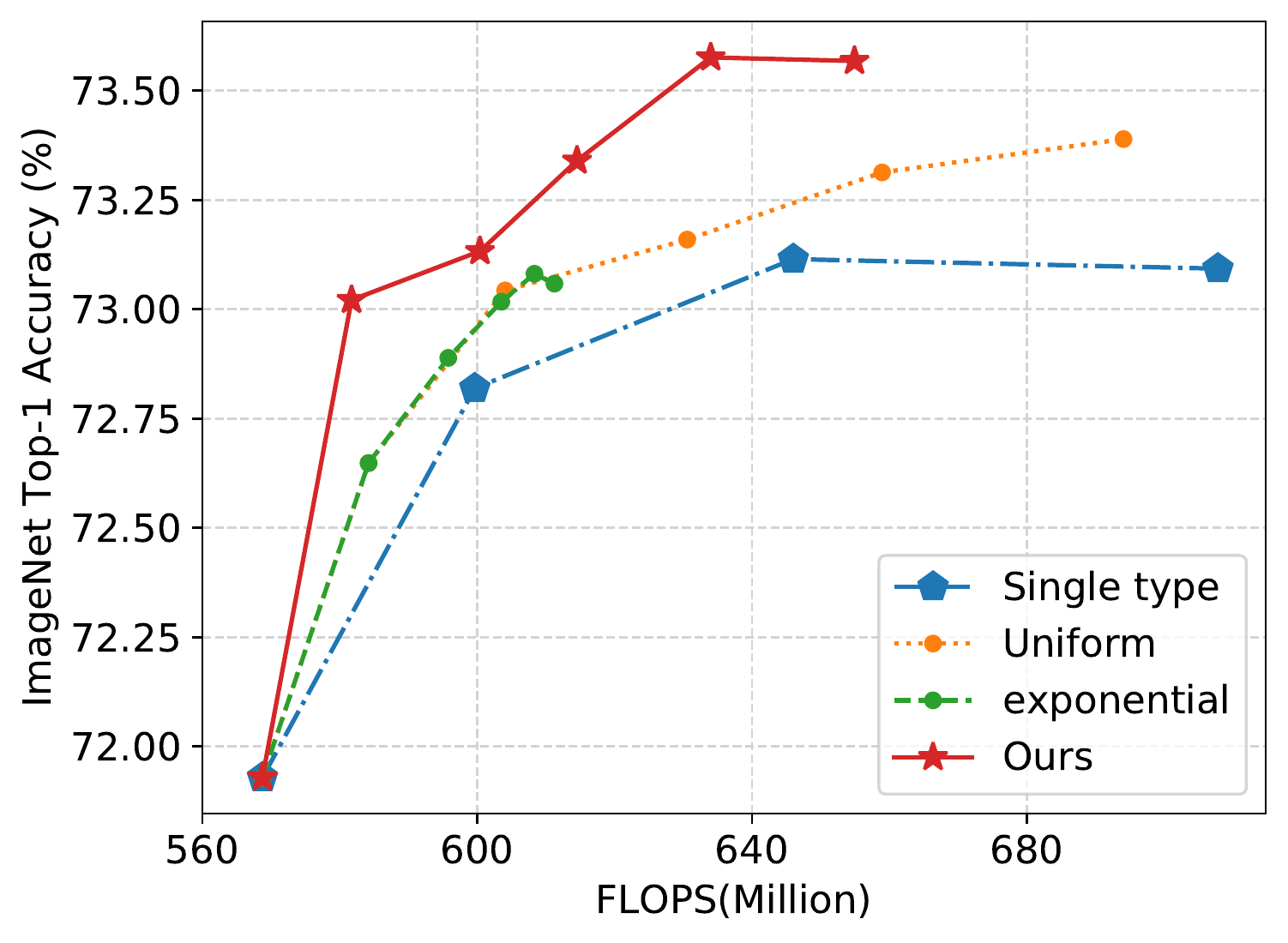}
\end{center}
   \caption{Comparison results on ImageNet. The searched model from our method achieves higher accuracy than both vanilla depthwise conv and the MixConv~\cite{tan2019mixconv} under the same FLOPs requirements. 
   }
\label{fig:mobilenetv1}
\end{figure}

\subsubsection{Results}
The experimental results are illustrated in Figure~\ref{fig:mobilenetv1}. For baseline A, similar to~\cite{tan2019mixconv}, we find that the model top-1 accuracy goes up when enlarging kernel size from $3\times3$ to $7\times7$ but starts dropping when kernel size is larger than $9\times9$. This can be explained that when kernel size is equal to the input feature in the extreme case, the convolutional layer simply becomes a fully-connected network, which is known to be harmful for performance~\cite{huang2017multi, tan2019mixconv}.

For baselines B and C, we observe that the depthwise convolution with multiple kernels   achieves  better FLOPs-accuracy trade-off than vanilla depthwise convolution and the performances of baseline-B and C are similar under the same FLOPs.  Besides, baseline B can be seen as a special case of uniform sampling.

Furthermore, with the same kernel candidate set,  our discovered models outperform both uniform and exponential allocation of kernels of different sizes,  under the same FLOPs constraint. We regard the performance gain is from the finer granularity of our search approach, which can  choose a suitable ratio of kernel sizes at different depth of the architecture. 

\subsection{Comparing with SOTAs}

To further demonstrate effectiveness of our search method, we compare  it  with state-of-the-art NAS methods.

\subsubsection{Settings}
Following ~\cite{wu2019fbnet, stamoulis2019single, cai2018proxylessnas}, we adopt the inverted residual bottleneck~\cite{sandler2018mobilenetv2} (MBConv) as our macro structure. The MBConv block is a sequence of a $1\times1$ pointwise convolution, $k \times k$ depthwise convolution, $1\times1$ convolution. Different from previous works that search for a single kernel size $k\times k$ in a depthwise convolution layer, our method   searches multiple kernel sizes.

The recent  MixNet~\cite{tan2019mixconv} also proposes to search among the mixed kernels. However, the ratio of different kernel sizes in their search space is fixed as uniform. In our search space, there is no constraints for the ratio of different kernel sizes so the search space is further enlarged. The search space in MixNet~\cite{tan2019mixconv} is a subset to  ours. The experimental results also show that our searched models achieve better performance-cost trade-off than MixNet.

\subsubsection{FLOPs vs. Accuracy}

\begin{figure}
    \centering
    \includegraphics[width=0.9\linewidth]{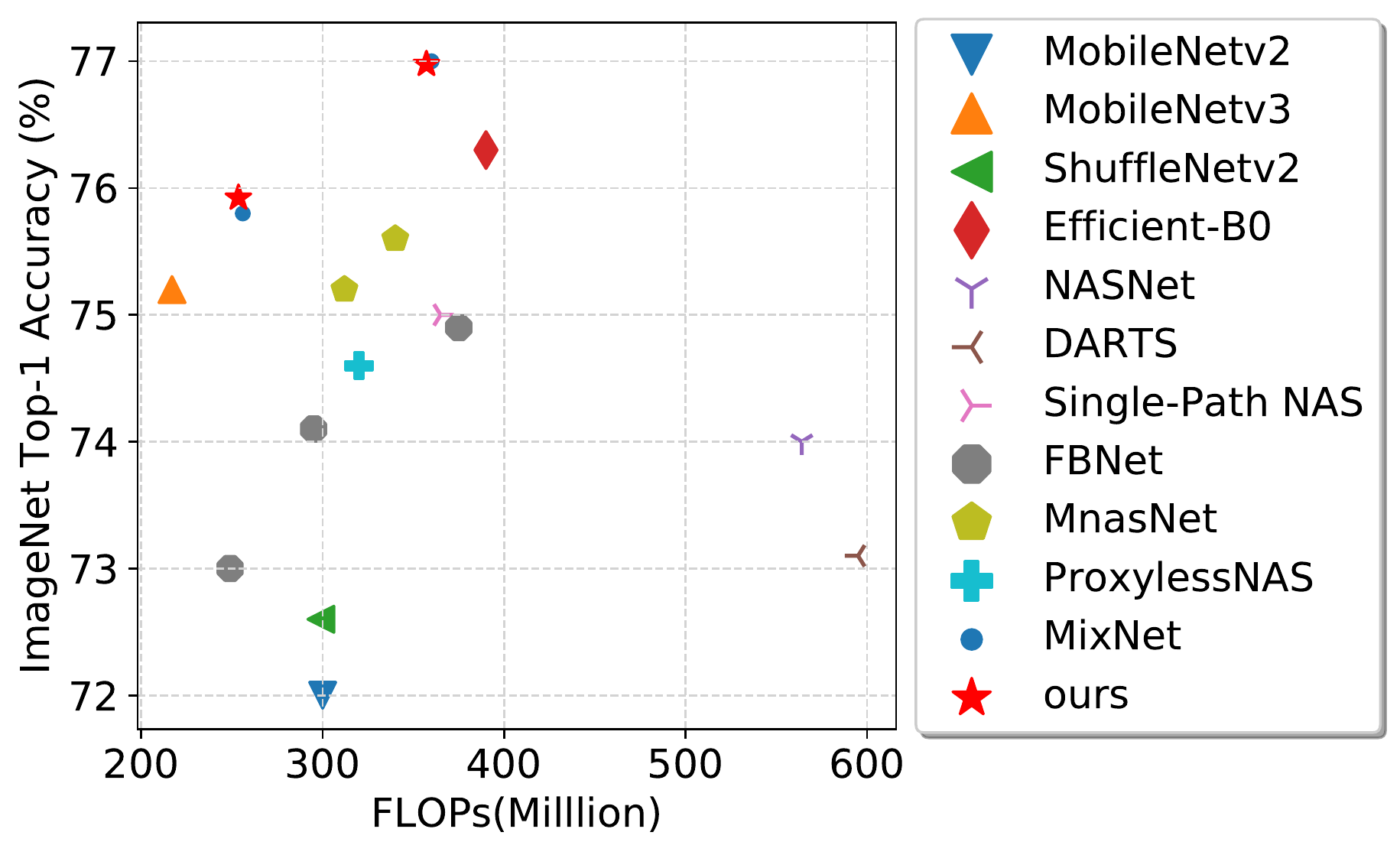}
    \caption{FLOPs versus Top-1 accuracy on ImageNet.}
    \label{fig:ImageNet-results}
\end{figure}

Evaluation results of our proposed \methodnameshort{} and comparison with state-of-the-art approaches are summarized in Table~\ref{tab:imagenet-results} and Figure~\ref{fig:ImageNet-results}. The \methodnameshort{}-A and \methodnameshort{}-B are obtained by setting different resource target 
, $\mathbi{T}$ in Eq.~\eqref{eq:flops-loss}. We set our target value as 260M, 370M respectively, which are set around the FLOPs of state-of-the-art model MixNet~\cite{tan2019mixconv} intentionally for  fair comparison under the similar FLOPs. 

As shown in the Table.~\ref{tab:imagenet-results}, our \methodnameshort{}-A achieves 75.9\% Top-1/92.9\% Top-5 accuracy with 254M FLOPs and \methodnameshort{}-B achieves 77.0\% Top-1/93.4\% Top-5 accuracy with 357 FLOPs. They outperform state-of-the-art manually designed models by a large margin. Specifically, our \methodnameshort{}-A is better than MobileNetV2 (+3.8\%) and ShuffleNetV2 (+3.2\%), with less FLOPs. 

Compared to recently proposed automated models generated by NAS methods, our \methodnameshort{} models perform better under similar FLOPs.  Specifically, compared to RL based methods, our \methodnameshort{}-A achieves 0.7\% higher Top-1 accuracy than MnasNet-A1~\cite{tan2019mnasnet} with 58M less FLOPs; 0.3\% higher Top-1 accuracy than MnasNet-A2 with 86M less FLOPs; 1.2\% higher Top1-accuracy than ProxylessNAS-R~\cite{cai2018proxylessnas} with 66M less FLOPs. Compared With gradient-based methods, \methodnameshort{}-A is better than ProxylessNAS-G (+1.6\%), Single-Path NAS~\cite{stamoulis2019single} (+0.8\%), FBNet-A/B/C (+2.8\%/+1.7\%/+0.9\%), respectively. 

\subsubsection{Searching Hours vs. Accuracy}
The comparison Results of GPU hours used for searching process is illustrated in Figure~\ref{fig:acc-GPUs}. Our search method is faster than most of the   methods   by a large margin. Compared to the recent state-of-the-art  model MixNet~\cite{tan2019mixconv} that also uses  multi-scale representation, our model \methodnameshort{}-A achieves a slightly higher accuracy (+0.1\%) than MixNet-S and the \methodnameshort{}-B achieves same accuracy as MixNet-M, while costing 3M less FLOPs. Remarkably, for achieving very similar results, our search method   needs about   GPU hours $1000 \times $ less than MixNet.

As mentioned in~\cite{wu2019fbnet, cai2018proxylessnas}, MnasNet~\cite{tan2019mnasnet} does not report the exact GPU hours for searching stage. In this work we adopt the the search cost data of MnasNet estimated in ProxylessNAS~\cite{cai2018proxylessnas}. For MobileNetv3~\cite{howard2019searching} and MixNet~\cite{tan2019mixconv}, as they use the same search framework as MnasNet, we roughly estimate  the search cost  of MobileNetv3 and MixNet to be similar to that of MnasNet.

\begin{table*}[h]
\begin{center}
\begin{tabular}{l|cccc|cc|cc}
\hline
Model  & \begin{tabular}[c]{@{}c@{}}Search \\ method\end{tabular} & \begin{tabular}[c]{@{}c@{}}Search \\ space\end{tabular} & \begin{tabular}[c]{@{}c@{}}Search\\ Dataset \end{tabular}   & \begin{tabular}[c]{@{}c@{}} GPU \\hours\end{tabular} & \#Params & \#FLOPs & \begin{tabular}[c]{@{}c@{}}Top-1 \\ acc (\%)\end{tabular} &  \begin{tabular}[c]{@{}c@{}}Top-5 \\ acc (\%)\end{tabular} \\ \hline
MobileNetV2~\cite{sandler2018mobilenetv2}           & manual    & -           &   -          & -                   & 3.4M & 300M &  72.0 & 91.0 \\
MobileNetV2($1.4\times$)                            & manual    & -           &   -          & -                   & 6.9M & 585M &  74.7 & 92.5 \\
ShuffleNetV2($1.5\times$)~\cite{ma2018shufflenet}   & manual    & -           &   -          & -                   & 3.5M & 299M &  72.6 &  -   \\
CondenseNet(G=C=4)~\cite{huang2018condensenet}      & manual    & -           &   -          & -                   & 2.9M & 274M &  71.0 & 90.0 \\
CondenseNet(G=C=8)                                  & manual    & -           &   -          & -                   & 4.8M & 529M &  73.8 & 91.7 \\ 
EfficientNet-B0~\cite{tan2019efficientnet}          & manual     &     -    & -              & -                    &  5.3M & 390M & 76.3  & 93.2 \\
\hline 
NASNet-A  \cite{zoph2018learning}                   & RL        & cell        &   CIFAR-10   & 48K                 & 5.3M & 564M &  74.0 & 91.6 \\
PNASNet \cite{liu2018progressive}                   & SMBO      & cell        &   CIFAR-10   & 6K$^{\dag}$         & 5.1M & 588M &  74.2 & 91.9 \\ 
AmoebaNet-A~\cite{real2019regularized}              & evolution & cell        &   CIFAR-10   & 75k                 & 5.1M & 555M &  74.5 & 92.0 \\
DARTS~\cite{liu2018darts}                           & gradient  & cell        &   CIFAR-10   & 96                  & 4.7M & 574M &  73.3 & 91.3 \\ 
P-DARTS~\cite{chen2019progressive}                  & gradient  & cell        &   CIFAR-10   & 7.2                 & 4.9M & 557M &  75.6 & 92.6 \\
GDAS~\cite{dong2019searching}                       & gradient  & cell        &   CIFAR-10   & 4.08                & 4.4M & 497M &  72.5 & 90.9 \\
\hline 
MnasNet-A1~\cite{tan2019mnasnet}                    & RL        & stage-wise  &   ImageNet   & 40K$^*$             & 3.9M & 312M &  75.2 & 92.5 \\ 
MnasNet-A2                                          & RL        & stage-wise  &   ImageNet   & 40K$^*$             & 4.8M & 340M &  75.6 & 92.7 \\ 
Single-Path NAS~\cite{stamoulis2019single}          & gradient  & layer-wise  &   ImageNet   & 30$^\curlywedge$             &  4.3M   & 365M    &  75.0 & 92.2 \\
ProxylessNAS-R~\cite{cai2018proxylessnas}             & RL        & layer-wise  &   ImageNet   & 200                 & 4.1M & 320M &  74.6 & 92.2 \\
ProxylessNAS-G                                        & gradient        & layer-wise  &   ImageNet   & 200                 &  -  & - &  74.2 & 91.7 \\
FBNet-A~\cite{wu2019fbnet}                          & gradient  & layer-wise  &   ImageNet   & 216                 & 4.3M & 249M &  73.0 & - \\  
FBNet-B~\cite{wu2019fbnet}                          & gradient  & layer-wise  &   ImageNet   & 216                 & 4.5M & 295M &  74.1 & - \\ 
FBNet-C~\cite{wu2019fbnet}                          & gradient  & layer-wise  &   ImageNet   & 216                 & 5.5M & 375M &  74.9 & - \\
MobileNetV3-Large~\cite{howard2019searching}        & RL        & stage-wise  &   ImageNet   &  40K$^\ddagger$      & 5.4M & 219M &  75.2 & - \\
MobileNetV3-Large($1.25\times$)        & RL        & stage-wise  &   ImageNet   &  -      & 7.5M & 356M &  76.2 & - \\
MobileNetV3-Small                                   & RL        & stage-wise  &   ImageNet   &  40K$^\ddagger$         & 2.9M & 66M  &  67.4 & - \\
MixNet-S~\cite{tan2019mixconv}                      & RL        & kernel-wise &   ImageNet   &  40K$^\ddagger$         & 4.1M & 256M &  75.8 & 92.8 \\
MixNet-M                                            & RL        & kernel-wise &   ImageNet   &  40K$^\ddagger$         & 5.0M & 360M &  77.0 & 93.3 \\
\hline 
\methodnameshort{}-A(ours)                          & gradient  & kernel-wise &   ImageNet   & \textbf{40}                  & 5.8M & 254M &  \textbf{75.9} & \textbf{92.9} \\
\methodnameshort{}-B(ours)                          & gradient  & kernel-wise &   ImageNet   & \textbf{40}                  & 7.2M & 357M &  \textbf{77.0} & \textbf{93.4} \\
\toprule  
\end{tabular}
\end{center}
\caption{ Performance of and \methodnameshort{} and state-of-the-art baseline architectures on ImageNet. For baseline models, we directly cite the parameter size, FLOPs, Top-1, Top-5 accuracy on the ImageNet validation set from their original papers. The data of search cost without additional superscript is also obtained from the original paper, while the superscripts means: * The search cost for MnasNet is obtained from ~\cite{cai2018proxylessnas}, in which Han \etal tested on V100 GPUs with the configuration described in ~\cite{tan2019mnasnet}. $\ddagger$ Both MobileNetv3~\cite{howard2019searching} and MixNet~\cite{tan2019mixconv} use the same search framework as MnasNet~\cite{tan2019mnasnet} so the search cost is roughly estimated based on MnasNet~\cite{tan2019mnasnet}.  $\curlywedge$ denotes TPU hours. $\dag$ The data is estimated in ~\cite{wu2019fbnet}.}
\label{tab:imagenet-results}
\end{table*}

\begin{table*}[!bt]
 \scriptsize
 \begin{center}
 \begin{tabular}{l |c | c | p{.26cm} p{.26cm} p{.26cm} p{.26cm} p{.26cm} p{.26cm} p{.26cm} p{.26cm} p{.26cm} p{.26cm} p{.26cm} p{.26cm} p{.26cm} p{.26cm} p{.26cm} p{.26cm} p{.26cm} p{.26cm} p{.26cm} p{.26cm}  }
  \toprule
  backbone  & FLOPs  & mAP       & aero      & bike      & bird      & boat      & bottle    & bus       & car       & cat       & chair     & cow       & table     & dog       & horse     & mbike     & persn     & plant     & sheep     & sofa      & train     & tv        \\ 
  \toprule
MBv1       & 10.16G & 75.9  &  83.9   &  79.3   &  75.1   &  65.8   &  55.9   &  84.3   &  85.7   &  85.4   &  58.4   &  80.9   &  70.4   &  82.0   &  84.9   &  84.5   &  79.7   &  48.3   &  77.8   &  76.6   &  84.1   &  74.3  \\
MBv2       & 9.10G  & 75.8 &  84.5   &  83.4   &  76.1   &  68.3   &  58.7   &  78.9   &  84.8   &  86.5   &  54.4   &  80.7   &  70.9   &  84.0   &  85.0   &  83.6   &  76.8   &  48.7   &  78.7   &  72.8   &  85.0   &  73.7   \\
MBv3-Small & 8.19G  &  69.3 &  77.4   &  76.9   &  67.0   &  62.0   &  43.7   &  76.3   &  79.1   &  82.1   &  47.2   &  75.4   &  65.2   &  78.4   &  81.0   &  79.7   &  72.5   &  39.4   &  68.7   &  67.1   &  76.1   &  70.0   \\
MBv3-Large & 8.78G  & 76.7  &  84.1   &  84.1   &  77.0   &  69.9   &  57.9   &  84.8   &  85.1   &  88.1   &  56.3   &  84.8   &  64.8   &  84.3   &  87.9   &  84.7   &  77.0   &  46.3   &  80.5   &  73.9   &  86.2   &  76.8  \\
MBv3-Large$^\dagger$ & 9.28G & 77.7 &  86.1   &  84.1   &  76.8   &  71.6   &  60.9   &  85.6   &  86.8   &  88.8   &  55.6   &  84.2   &  70.2   &  85.7   &  86.2   &  85.0   &  77.6   &  47.5   &  81.6   &  73.5   &  88.1   &  77.6  \\
\hline
\methodnameshort{}-tiny   &   8.73G & 76.2 &  79.6   &  83.5   &  76.4   &  68.1   &  54.3   &  83.2   &  85.6   &  87.1   &  56.0   &  82.8   &  71.9   &  84.9   &  85.0   &  85.7   &  76.4   &  47.1   &  81.7   &  75.0   &  85.5   &  74.7 \\  
\methodnameshort{}-A &   8.91G &  77.3 &  86.1   &  84.8   &  76.8   &  68.6   &  59.2   &  83.6   &  86.3   &  87.1   &  56.9   &  85.2   &  67.2   &  86.6   &  87.2   &  86.0   &  77.7   &  49.1   &  80.8   &  74.5   &  86.3   &  76.7 \\
\methodnameshort{}-B &  9.27G  &  78.0   &  86.8   &  84.5   &  77.5   &  69.8   &  58.2   &  85.3   &  86.4   &  88.3   &  60.4   &  84.5   &  72.9   &  85.8   &  86.7   &  86.8   &  78.0   &  51.3   &  80.0   &  73.9   &  86.4   &  77.4  \\
\toprule
\end{tabular}
\end{center}
\caption{Results on VOC2007 \texttt{test}. $\dagger$ denotes MobileNetV3 with the multiplier=1.25. Note that here the FLOPs of the total detection network is dominated by the detection header.}
\label{tab:voc2007test}
\end{table*}

\subsection{Object Detection}
To further validate the effectiveness of our the \methodnameshort{} models, we conduct object detection experiments on the PascalVOC~\cite{everingham2010pascal} dataset. Following the broadly used strategy, we combine the VOC2007 \texttt{trainval} and VOC2012 \texttt{trainval} as the training data and test the performance of our model on VOC2007 \texttt{test}. We adopt our \methodnameshort{} as a drop-in backbone feature extractor in YOLOV3~\cite{redmon2018yolov3}. All backbone models are pre-trained on ImageNet and fine-tuned on PascalVOC for 200 epochs. 

We first train the model with $\text{lr}=10^{-3}$  for 160 epochs, and then continue the training with $\text{lr}=10^{-4}$ for 20 epochs and $\text{lr}=10^{-5}$ for another 20 epochs. The results on VOC2007 \texttt{test} are shown in  Table~\ref{tab:voc2007test}. Our \methodnameshort{}-A model outperforms the MobileNetv1~\cite{howard2017mobilenets}, MobileNetv2~\cite{sandler2018mobilenetv2} feature extractors by 0.3 mAP, 0.4 mAP, respectively, while consuming less FLOPs. And our \methodnameshort{}-B performs better than MobileNetV3($\times1.25$) that has a nearly same FLOPs as our model.

\subsection{Visualization on Kernel Size Distribution}

\begin{figure}
\begin{center}
\includegraphics[width=0.9\linewidth]{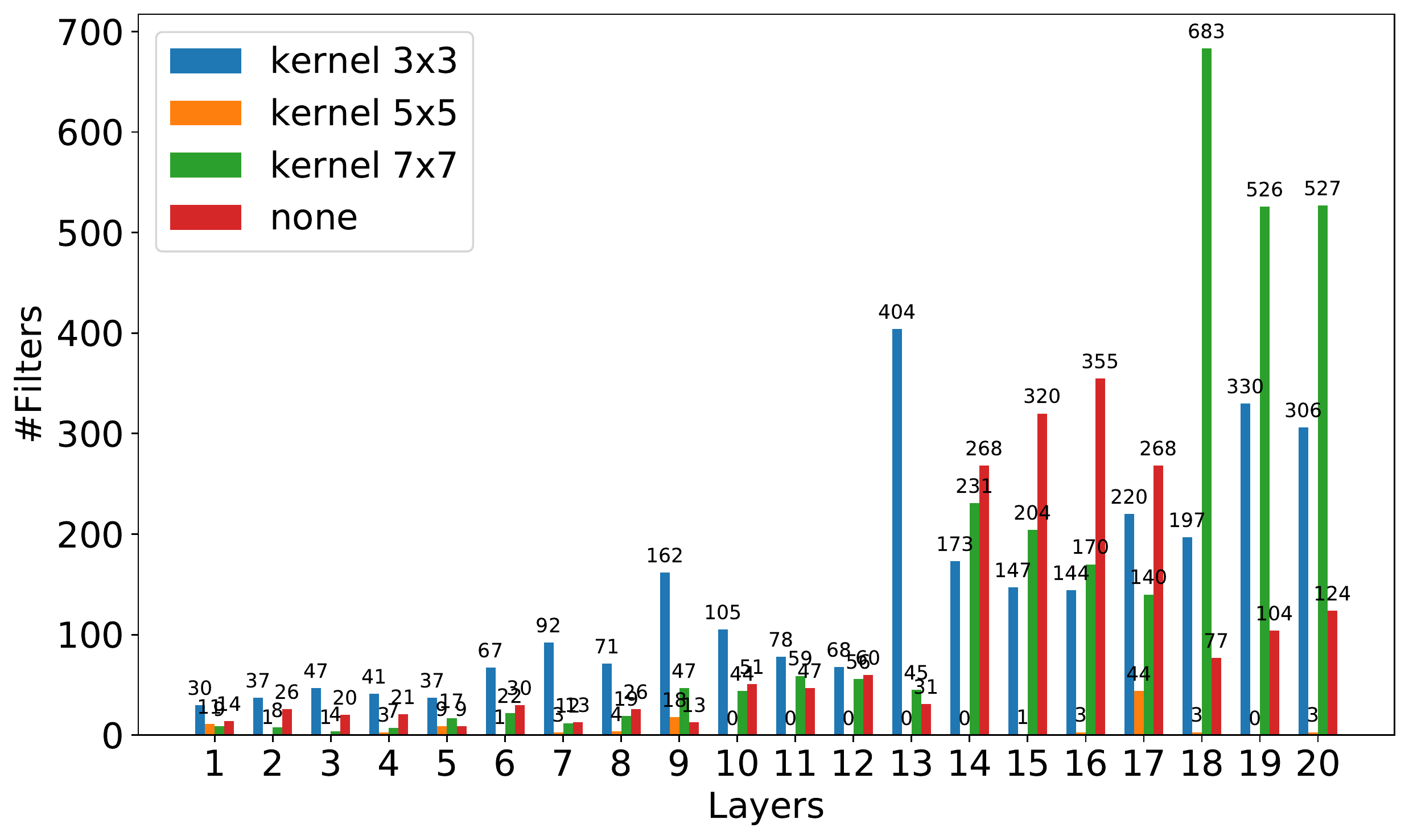}
\end{center}
   \caption{Illustration of the distribution of different kernel sizes across the network. The small kernels $3\times3$ account for a large proportion in the shadow layer, and lager kernels are preferred in the deep layer.}
\label{fig:kernelsdis}
\end{figure}

As our search algorithm can determine the number of different kernel sizes automatically, we wonder what is  the intrinsic preference on kernel sizes for CNNs. We plot the distribution of each kernel size in Figure~\ref{fig:kernelsdis}. We observe   that at the shallow layers, the network tends to choose   smaller kernel size. With the layer going deeper, large kernels begin to  occupy a larger proportion. Interestingly, our findings are consistent with the MixConv~\cite{tan2019mixconv}.  This interesting findings may inspire future works for understanding CNN. 

%% file: tex/6_discussion.tex
\section{Discussion}
In this work,  we treat every kernel within a single depthwise convolution independently and search for a mixed convolution. In this way, our proposed search strategy can search in a more fine-grained way. 
Furthermore, our method is also compatible with atrous convolution~\cite{chen2017rethinking}, asymmetric convolution~\cite{jin2014flattened, szegedy2016rethinking, tai2015convolutional} following the same rule as discussed in Sec~\ref{sec:mask-step}, not requiring any other adaptation. So, our method can be equipped to existing works~\cite{liu2018darts, xie2018snas, dong2019searching} to further improve searching efficiency.

%% file: tex/7_conclusion.tex
\section{Conclusion}
In this work, we propose an efficient search strategy to reduce the search cost dramatically. We encode the supernet from the perspective on convolution kernels rather than on feature maps, which could drop the requirement for memory and computation resources remarkably. Specifically, our search process is about faster than MnasNet by three orders of magnitude. Our proposed method digs deep into the more fine-grained search space, i.e, convolutional kernels. We demonstrate experimentally that our discovered models achieve better performance on ImageNet under the same computation resource constraints. We hope that our research will be beneficial in accelerating the search procedure and further promote the development of NAS.